\documentclass[conference]{IEEEtran}
\IEEEoverridecommandlockouts
\usepackage{cite}
\usepackage{amsmath,amssymb,amsfonts}
\usepackage{algorithmic}
\usepackage{graphicx}
\usepackage{textcomp}
\usepackage{xcolor}
\usepackage{listings}
\usepackage[format=plain, justification=justified, labelfont=small, textfont=small ]{caption}
\renewcommand{\thetable}{\Roman{table}}
\usepackage{transparent}
\usepackage{tabularx}

\definecolor{LightGray}{rgb}{0.97,0.97,0.97}
\definecolor{Sepia}{rgb}{0.60,0.37,0.17}
\definecolor{brick}{rgb}{0.73,0.13,0.13}
\definecolor{grey}{rgb}{0.97,0.97,0.97}
\definecolor{blue}{rgb}{0,0,1}
\definecolor{black}{rgb}{0,0,0}
\definecolor{green}{rgb}{0,0.5,0}
\definecolor{Purple}{rgb}{0.63,0.13,0.94}
\definecolor{MidnightBlue}{rgb}{0.60,0.37,0.17}
\lstdefinelanguage{SPARQL}{
  basicstyle=\footnotesize\ttfamily,
  columns=fullflexible,
  breaklines=false,
  sensitive=true,
  xleftmargin=.5em,
  xrightmargin=.5em,
  framexleftmargin=.5em,
  framextopmargin=.5em,
  framexbottommargin=.5em,
  framexrightmargin=.5em,
  tabsize = 2,
  showstringspaces=false,
  morecomment=[l][\color{gray}]{\#},       
  morecomment=[n][\color{blue}]{<http}{>}, 
  morestring=[b][\color{brick}]{\"},  
  keywordsprefix=?,
  classoffset=0,
  keywordstyle=\color{black},
  morekeywords={},
  classoffset=1,
  keywordstyle=\color{blue},
  morekeywords={rdf,rdfs,owl,xsd,purl,OpcUa,OpcSS,VDI3682,DINEN61360,ISA88},
  classoffset=2,
  keywordstyle=\color{green},
  morekeywords={
    SELECT,CONSTRUCT,DESCRIBE,ASK,WHERE,FROM,NAMED,PREFIX,BASE,OPTIONAL,FILTER,GRAPH,LIMIT,OFFSET,SERVICE,UNION,EXISTS,NOT,BINDINGS,MINUS,a, ORDER, BY, ASC, DESC, 
  }
}

\def\BibTeX{{\rm B\kern-.05em{\sc i\kern-.025em b}\kern-.08em
    T\kern-.1667em\lower.7ex\hbox{E}\kern-.125emX}}

\begin{document}

\title{Accessing and Interpreting OPC UA Event Traces based on Semantic Process Descriptions\\
}

\author{\IEEEauthorblockN{Tom Westermann}
\IEEEauthorblockA{\textit{Institute of
Automation Technology},\\
\textit{Helmut-Schmidt-University}\\
Hamburg, Germany \\
tom.westermann@hsu-hh.de}
\and
\IEEEauthorblockN{Nemanja Hranisavljevic}
\IEEEauthorblockA{\textit{Fraunhofer Center for Machine Learning} \\
\textit{Fraunhofer IOSB-INA}\\
Lemgo, Germany \\
nemanja.hranisavljevic@iosb-ina.fraunhofer.de}
\and
\IEEEauthorblockN{Alexander Fay}
\IEEEauthorblockA{\textit{Institute of
Automation Technology},\\
\textit{Helmut-Schmidt-University}\\
Hamburg, Germany \\
alexander.fay@hsu-hh.de}
}

\maketitle

\begin{abstract}
The analysis of event data from production systems is the basis for many applications associated with Industry 4.0. 
However, heterogeneous and disjoint data is common in this domain.
As a consequence, contextual information of an event might be incomplete or improperly interpreted which results in suboptimal analysis results.
This paper proposes an approach to access a production systems' event data based on the event data's context (such as the product type, process type or process parameters). 
The approach extracts filtered event logs from a database system by combining: 1) a semantic model of a production system's hierarchical structure, 2) a formalized process description and 3) an OPC UA information model. 
As a proof of concept we demonstrate our approach using a sample server based on OPC UA for Machinery Companion Specifications.
\end{abstract}


\section{Introduction} \label{Introduction}
One vision of Industry 4.0 are cyber-physical production systems (CPPS) that are self-diagnosing, self-adapting and self-optimizing \cite{A.Bunte.2019}. During production, these systems generate large amounts of multivariate data, which contain information about the systems' behaviour. 
In many applications, such as bottleneck analysis, predictive maintenance or anomaly detection, the stored data can be used to learn a model which captures the systems' behaviour. 
Often, these models take the form of Petri nets, Markov chains or various types of automata\cite{A.Maier.2015}\cite{J.Ladiges.2015}.

In practice, time-annotated production data is difficult to process\cite{VogelHeuser.2021} due to numerous reasons. 
The system architecture of production systems is heterogeneous and consists of many different interacting systems whose components 
are produced by different manufacturers and may use heterogeneous data models \cite{Jirkovsky.2017}. 
These components often have their own data storage, which could be relational databases, log files or other formats. 
Across these different data silos, 
similar concepts might be represented in different ways.
This heterogeneous and complex character of the data is one of the biggest obstacles for the exchange of knowledge in automation systems \cite{Schneider.2019} and big data applications in general \cite{Yin.2015}.

\begin{figure}[t] 
\centering 
\includegraphics[scale=1]{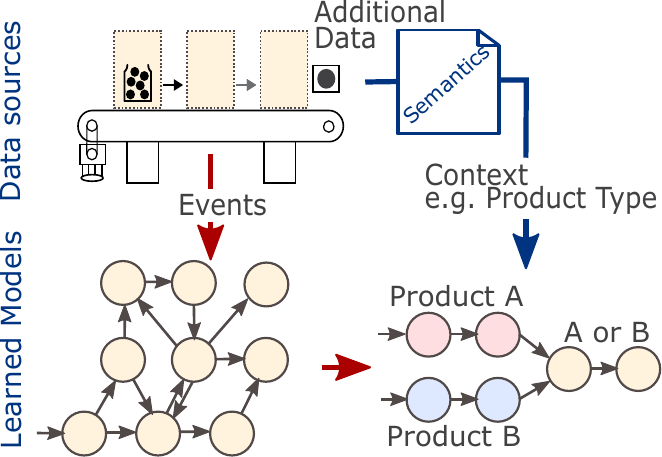} 
\vglue 0ex plus 0.5ex minus 3.5ex 
{\caption{\small Semantically enriching event logs with context and additional data may result in more accurate and understandable analysis results. }\label{motivation}}%
\end{figure}

Another problem encountered during the analysis of event traces is the unavailability of prior knowledge. 
In analysis, this lack of information can cause "rediscovering" information that is already known but unavailable, or lead to wrong conclusions about the system's behaviour.
This is depicted in Figure \ref{motivation}.
It shows two models, one learned from the raw event stream, and the other learned from raw events and additional knowledge like the type of process and the type of product that was produced. The first model might be complex and opaque to the operator. 
However, it is expected that enriching the 
logs with information from additional data sources can improve the learned model both in terms of accuracy as well as interpretability. 

Semantic Web Technologies (SWT) like ontologies, the query language \textit{SPARQL} or  knowledge graphs can provide a solution to the described problems \cite{Xiao.2019}. 
An ontology can provide an explicit specification
of the concepts and their relationships in a domain. If this conceptualization is combined with instances and their relations, this can be considered a knowledge graph. 
This knowledge graph can provide a way to align incompatible data that semantically refers to the same entity. With ongoing research in the field of ontology-based data access (OBDA), Semantic Web Technologies can also be used to integrate information from distinct data silos.  
With regard to the analysis of time-annotated event traces from CPPS, knowledge graphs also offer a way to provide formalized prior knowledge, i.e. a machine interpretable model of implicit knowledge that is otherwise not available in the analysis. 

This paper proposes a knowledge graph that allows extracting time spans that correspond to the execution of production processes at a CPPS, based on a formalized process description. 

The extracted process time spans can then be used along with an OPC UA information model to access a time-series database. This makes it possible to retrieve event traces that correspond to these processes. 

The remainder of the paper is structured as follows: Section \ref{relatedWorks} presents and discusses relevant research. Section \ref{semanticModel} defines an ontology to access variable values from an OPC UA information model based on a formalized process description.  
A prototypical implementation of this is shown in Section \ref{useCase} and evaluated in Section \ref{validation}.

\section{Related Work} \label{relatedWorks}
This section gives an overview of numerous works that used Semantic Web Technologies to either model information about sensors of a technical resource, or access this type of information through an OPC UA server.

Jiskovsky et al. \cite{Jirkovsky.2017} propose a semantic big data historian that uses Semantic Web Technologies to integrate sensor data from a hydroelectric power plant and combines it with additional information from other systems. It proves successful at resolving heterogeneity between different systems. However, the sensor measurements are stored inside the triple store, which can have a negative effect on the graph's size and performance. The knowledge graph does not consider possibilities for reuse. 

Schiekofer et al. provide a formal mapping between OPC UA and technologies of the semantic web stack \cite{Schiekofer.2019}. This allows for the explicit description of OPC UA semantics, which are usually only implicitly defined in the information models' documentation. The authors also offer methods for consistency checking of the OPC UA information model using a reasoner and querying the information model using SPARQL queries.

Hildebrandt et al. present a domain-expert-centric approach to ontology design of CPPS, focusing on 
industry standards as ontology design patterns \cite{Hildebrandt.2020}. The approach is validated on an industrial use case encompassing system structure, processes, states, and data elements and has a strong focus on ontology reuse. Access to high-velocity data that should not be stored in the ontology is not in the scope of the paper. 

Kalayc{\i} et al. use OBDA to access heterogeneous manufacturing data  from a knowledge graph \cite{GuzelKalayc.2020}. This knowledge graph represents the concepts and properties that are relevant for surface mounting process manufacturing, together with important domain knowledge. Using the OBDA-tool Ontop, SPARQL queries were executed to answer analytical queries about the production and failures. The approach is only applicable to relational databases, which can be accessed via SQL queries. 

Xiao et al. \cite{Xiao.2019} list further use cases of OBDA, spanning diverse domains including manufacturing, process mining and administration, while Ekaputra et al. \cite{Ekaputra2017} review OBDA approaches in multi-disciplinary engineering environments.

Steindl et al. propose a different ontology-based method to access OPC UA data through a regular SPARQL endpoint and Custom Property Functions (CPF). These CPF extend the SPARQL query evaluator of the Apache Jena Framework with custom code, that gets executed whenever the CPF gets called in a query. In order to avoid overloading the triple store, the OPC UA data is stored in a separate database and only accessed on-demand \cite{G.Steindl.2019a}. It succeeds in accessing the recorded values. Queries that rely on an additional knowledge outside the extracted OPC UA information model were not in the scope of the paper. This approach relies on OPC UA Historical Data Access, which accesses historical values through the OPC server itself. For read-only tasks, Mathias et al. recommend accessing the database directly, because passing through the OPC UA server might cause computational overhead on the PLC \cite{Mathias.2020}.

Steindl et al. further evaluated the query execution times of three different methods to integrate time series data into knowledge graphs \cite{G.Steindl.2019b}. The authors found that both Ontop and custom property functions were superior to storing the time series data inside the knowledge graph. If the time series data is already stored in a preexisting SQL database, they recommend Ontop. Otherwise they recommend using CPFs.

While there are many approaches to model static information about CPPS, no method of accessing segments of time-annotated sensor data based on this information exists.
This work therefore aims to close the gap between the approach to ontology modeling from Hildebrandt et al. \cite{Hildebrandt.2020} and the data access to logged OPC UA data from Steindl et al. \cite{G.Steindl.2019a}. The approach from Hildebrandt et al. \cite{Hildebrandt.2020} is used to formalize prior knowledge about the structure and processes of a CPPS. Based on this process description, process-based event traces can be extracted using the approach from Steindl et al. \cite{G.Steindl.2019a}.

\section{Ontology of Processes, Production Systems and their Variables} \label{semanticModel}
As described in Section \ref{Introduction}, the analysis of CPPS event traces is hindered by semantic heterogeneity across data sources, separate data silos and unavailability of prior knowledge during analysis. 
To overcome these issues, this section introduces an ontology that formalizes prior knowledge about processes and their types. It is then combined with additional information on events and time series that were recorded at a CPPS from an OPC UA server. 

The creation of the ontology follows the domain-expert-centric-approach to industrial ontology design outlined by Hildebrandt et al. \cite{Hildebrandt.2020}, in which suitable ontology design patterns (ODP) are identified based on user-defined competency questions (CQ) that the ontology is supposed to answer.  

\subsection{Requirements and Design Patterns}
The goal of the ontology is to provide semantic access to time-annotated data of CPPSs. Therefore, the first requirement is that the model needs access to the full event trace of a CPPS. This requirement can in turn be divided into two distinct competency questions. 

The first competency question asks for information of which variables belong to the CPPS in question. 
In modern CPPSs, this information is available in the OPC UA servers information model. This model describes standardized nodes and their relationships of a server’s address space. They can be modeled according to OPC UA companion specifications, which further specify models e.g. for specific industries.

Since queries should refer to individual machines, OPC UA for Machinery is a suitable companion specification to use in this work\cite{OPC40001-1}. It allows clear identification of individual machines and the variables they organize. Because OPC UA for Machinery functions as a container for machine representations that follow other companion specifications (e.g. machine tools or robotics), it is not restricted to a single domain.
All parts of the information model that are relevant for this use case can be translated into OWL. To achieve this, the OPC UA information models can be automatically extracted and transformed into RDF-triples \cite{Hildebrandt.2020} \cite{G.Steindl.2019a}.

\begin{table}[ht]
    \begin{tabular}{l l|l|l|l} \hline
        \rule{0pt}{2ex}RQ 1:& \multicolumn{4}{l}{Full Event trace of single machine} \\ 
        \hline
        \rule{0pt}{2ex}CQ 1.1: & \multicolumn{2}{p{3cm}}{Which variables belong to a certain machine?} & \multicolumn{2}{p{3cm}} {Machine x organizes variables a, b, c} \\ 
        CQ 1.2: & \multicolumn{2}{p{3.5cm}}{Which events and values were recorded at a specific machine in a specific timeframe?} & \multicolumn{2}{p{3cm}} {Events \textit{x, y, z} happened at Variables  \textit{a, b, c}} \\ \hline
        \rule{0pt}{2ex}ODP: & \multicolumn{4}{p{6.5cm}}{OPC UA for Machinery + Companion Specification} \\ \hline
    \end{tabular}
    \caption{Requirement and derived competency questions for RQ1.}
    \label{tab:RQ1}
\end{table}

The second competency question refers to the extraction of variable value changes that were recorded in a given time interval. This opens up the question of data storage and access. Since these values are recorded and written at short intervals, they are not well suited to be stored inside a graph database. Multiple approaches for access to this type of data exist in the literature, among them Virtual Knowledge Graphs\cite{GuzelKalayc.2020} or CPFs. For this use case a CPF was chosen following the approach from \cite{G.Steindl.2019a}. Further information on the implementation can be found in Section \ref{useCase}.

As described in Section \ref{Introduction}, similar production processes are expected to produce similar event traces during their execution. Therefore the access to these values should be based on the production process that was performed at the time the events occurred. 
To classify these production processes, an ODP based on the ISA-88 model for batch control can be used\cite{Isa88-1}. It contains three types of information: The physical model describes physical assets and their hierarchy, while the procedural control model describes abstract recipes and operations that these physical assets can perform. Additionally, it includes a process model that describes realizations of these abstract recipes and operations.

These process realizations can be described in greater detail using the Formalized Process Description according to the VDI 3682\cite{VDI/VDE3682-1}. 
It offers a formalized and easily understandable description mechanism. It can be used to describe which technical resource performed a process and which products/energy/information elements were used as inputs or outputs. Especially relevant to the use case in this paper would be an information element regarding the start and end timestamp of the process. Previous works that used this approach to ontology design modeled information elements according to the data element defined in DIN EN 61360 \cite{DINEN61360-1}\cite{C.Hildebrandt.2019}. The data element contains a type description of the element as well as an instance description, which can contain the value associated with the data element. 

\begin{table}[h]
\begin{center}
    \begin{tabular}{l l|l|l|l} \hline
        \rule{0pt}{2ex}RQ 2:& \multicolumn{4}{p{6.5cm}}{ Events and time series data of machines should be accessible based on their context} \\ 
        \hline
        \rule{0pt}{2ex}CQ 2.1: & \multicolumn{2}{p{3cm}}{Which events happened at a machine during a specific type of production process?} & \multicolumn{2}{p{3cm}} {Events \textit{x, y, z} happened at Variables  \textit{a, b, c}} \\ 
        CQ 2.2: & \multicolumn{2}{p{3.5cm}}{Which events happened at a machine during production of a certain product type?} & \multicolumn{2}{p{3cm}} {Events \textit{x, y, z} happened at Variables  \textit{a, b, c}} \\ \hline
        \rule{0pt}{2ex}ODP: & \multicolumn{4}{p{6.5cm}}{\raggedright ISA-88: Physical assets and process types \linebreak VDI3682: Description of process instances \linebreak DINEN61360: Data Elements} \\ \hline
    \end{tabular}
    \caption{Requirement and derived competency questions for RQ2.}
    \label{tab:RQ2}
\end{center}
\end{table}

\subsection{Lightweight Ontology}
\begin{figure}[htp]
    \centering
    \includegraphics[width=0.48\textwidth]{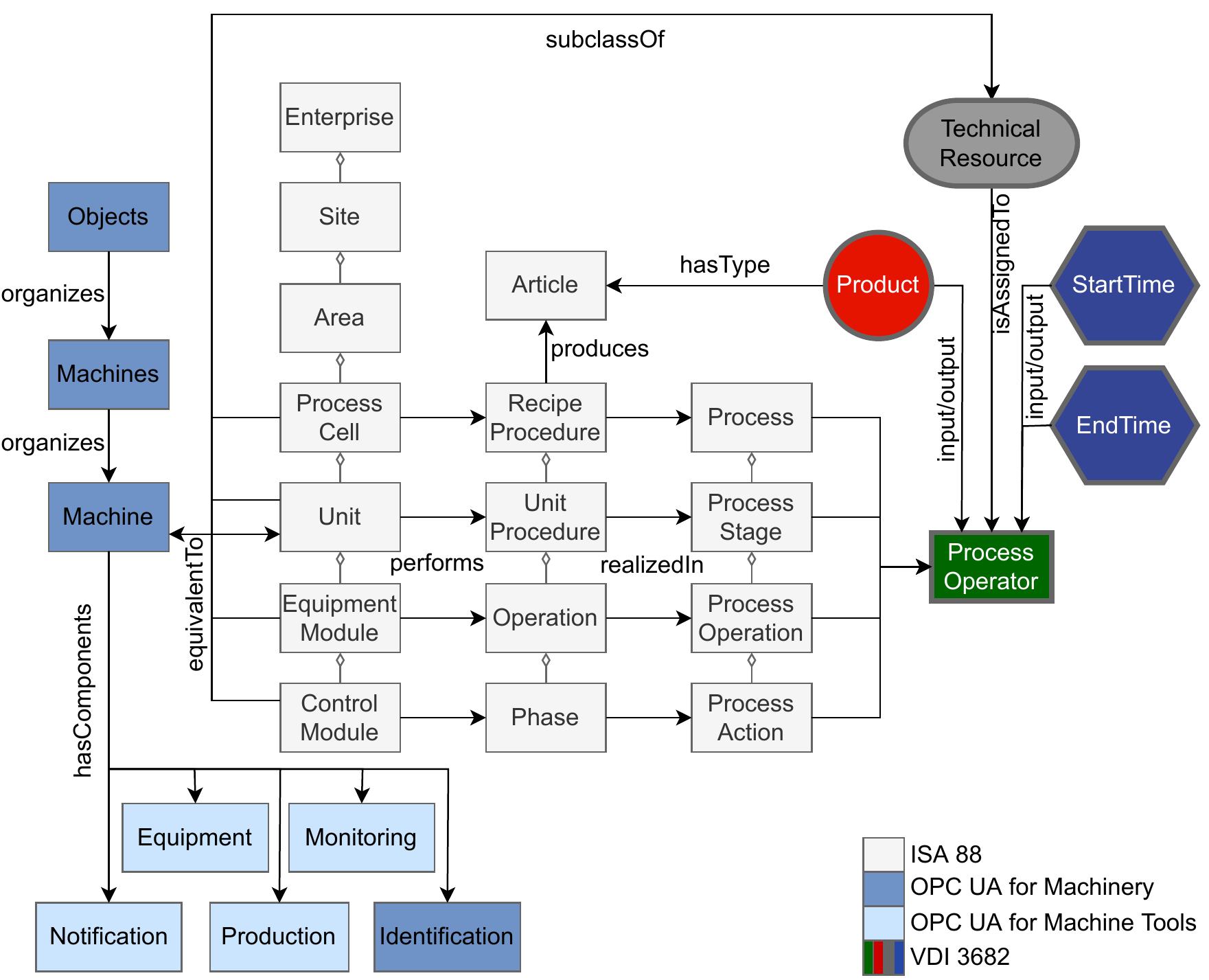}
    \caption{Lightweight ontology of the semantic integration of ISA 88, VDI3682 and OPC UA for Machinery. Please note that both OPC UA information models are very large and were therefore truncated.}
    \label{fig:LightweightOntology}
\end{figure}

The use of three different ODPs causes semantic heterogeneity between some concepts of the ontology that needs to be resolved. 
Between the ODPs of the ISA 88 and the VDI 3682, this heterogeneity is twofold. Any \textit{Physical Asset} in the ISA 88 model that can perform a part of a \textit{Recipe Procedure} can be considered a \textit{Technical Resource}. Therefore, they can be considered subclasses of the broader term \textit{Technical Resource} from the VDI 3682. Additionally, any \textit{Process} in the ISA 88 process model can be described as a \textit{Process Operator} from the VDI 3682. These are therefore subclasses of the \textit{Process Operator} as well. This is visualized in Figure \ref{fig:LightweightOntology}.

Further semantic heterogeneity exists between the OPC UA for Machinery information model and the physical model from the ISA 88. While the ISA 88  model describes a machine with regards to the kinds of operations it can perform, the OPC UA for Machinery model describes a machine along with the data that it generates. The machine itself should however be represented in the ISA 88 model as well. Since the machines in the OPC UA for Machinery model are individually capable of performing processes, they correspond to \textit{Units} in the ISA 88 model.

\section{Use Case: Process-Based access to event logs of an OPC UA Server} \label{useCase}

\begin{figure}[htp]
    \centering
    \includegraphics[width=0.4\textwidth]{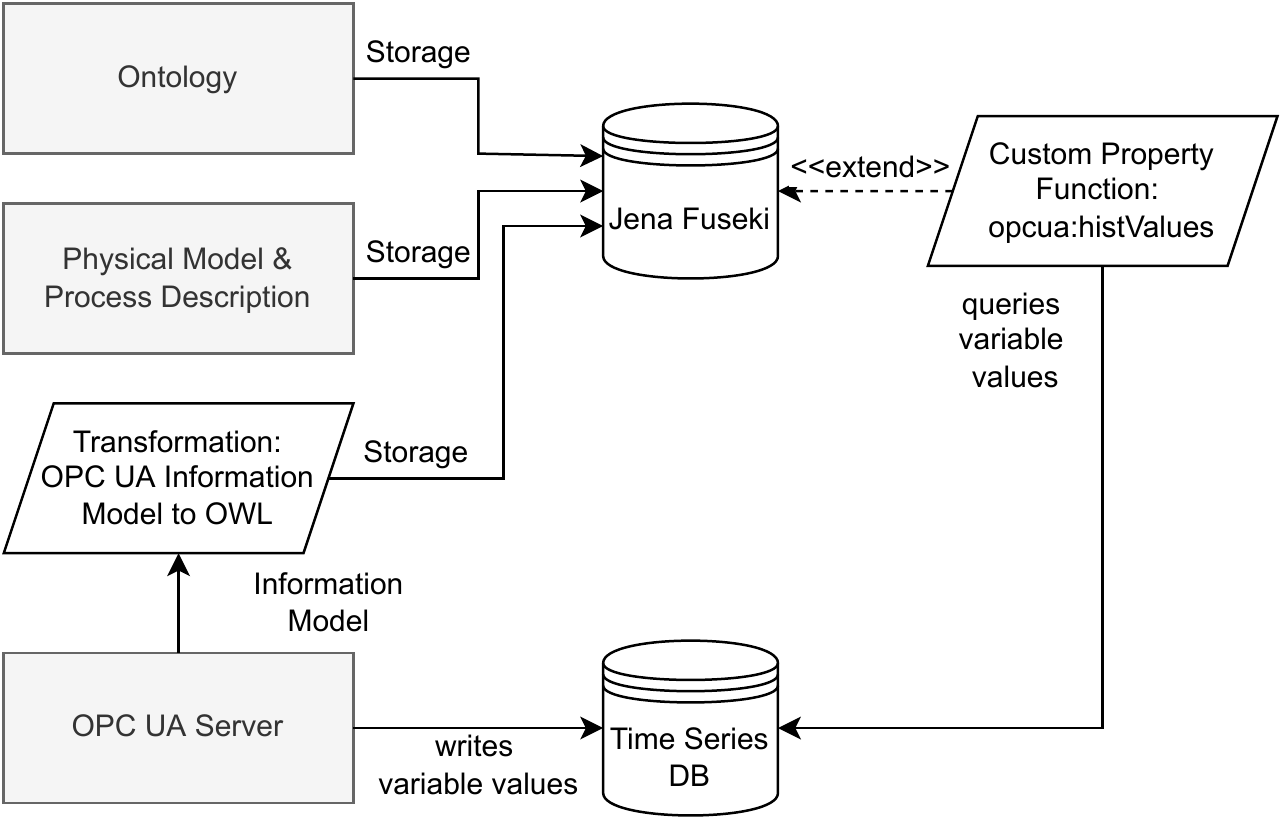}
    \caption{Storage and access setup that was used in the validation experiment. Ontology, physical model and process description were materialized along with the OWL-representation of the OPC UA information model.}
    \label{fig:DataStorage}
\end{figure}

In order to validate the functionality of the described approach, the ontology described in Section \ref{semanticModel} was implemented and filled with data describing a machine and process hierarchy according to ISA88. 
The knowledge about the production facility was modeled in OWL and stored in a knowledge graph. The graph was then stored in Apache Jena\footnote{https://jena.apache.org/}, which is an open source framework to build applications based on Semantic Web Technologies (s. Figure \ref{fig:DataStorage}).

For this validation, a simulated OPC UA sample server from UMATI was used \cite{christian_von_arnim_2022_6336935}. Its information model follows the OPC UA for Machinery companion specifications \cite{OPC40001-1}. Inside its \textit{Machines}-folder, it contains various machines (eg. machine tools\cite{OPC40501-1}, woodworking tools, robots etc.) that follow their respective OPC UA Companion Specifications. These machines were considered part of a physical model according to ISA88., i.e. a \textit{ISA88:Unit} that can perform \textit{ISA88:UnitProcedures}.
The OPC UA information model was extracted via an automated node crawler and transformed into OWL using the Open Source Tool \textit{Lion}\footnote{https://github.com/hsu-aut/lion}.

In order to log the events and time series data from the OPC UA server, multiple approaches exist. The different events and time series values can be materialized inside the triple store. Due to the high volume  and velocity of measurements however, this approach would result in inefficient query times \cite{G.Steindl.2019b} and data processing in general \cite{Jirkovsky.2017}.
Because of these limitations, a dedicated time series database was favored instead. For this task, InfluxDB\footnote{https://portal.influxdata.com/downloads/} was chosen since it is well suited for high write loads. All variable value changes from the machines on the OPC UA server were logged using the open source server agent \textit{Telegraf}\footnote{https://www.influxdata.com/time-series-platform/telegraf/} and its OPC UA plugin\footnote{https://www.influxdata.com/integration/opcua/}. The time series is stored in InfluxDB according to the variable's \textit{nodeId}, so that all information necessary for the query can be extracted from the ontology. This way, all value changes of variables that have an OPC UA type definition of \textit{BaseDataVariableType}, \textit{AnalogUnitRangeVariableType} or \textit{FiniteStateVariableType} are logged in the database.

To allow direct access from the ontology, the approach by Steindl et al. \cite{G.Steindl.2019a} was modified. As in \cite{G.Steindl.2019a}, a custom property function was registered as an extension in Apache Jena's SPARQL processor \textit{ARQ}. If executed, the custom property function queries the node's variable value changes from a database for any nodes that match the graph pattern specified in the rest of the SPARQL query. 
The historical values can therefore be accessed by calling the custom property function \textit{?node histValues(?time ?value ?starttime ?endtime)} with \textit{node} being the selected node from the OPC UA information model and \textit{histValues()} being the parametrized custom property function.
For the given case, a connector to InfluxDB was implemented that automatically queries a variables historical data between two timestamps using InfluxDB's query language.

\begin{figure}[htp]
    \centering
    \includegraphics[width=0.44\textwidth]{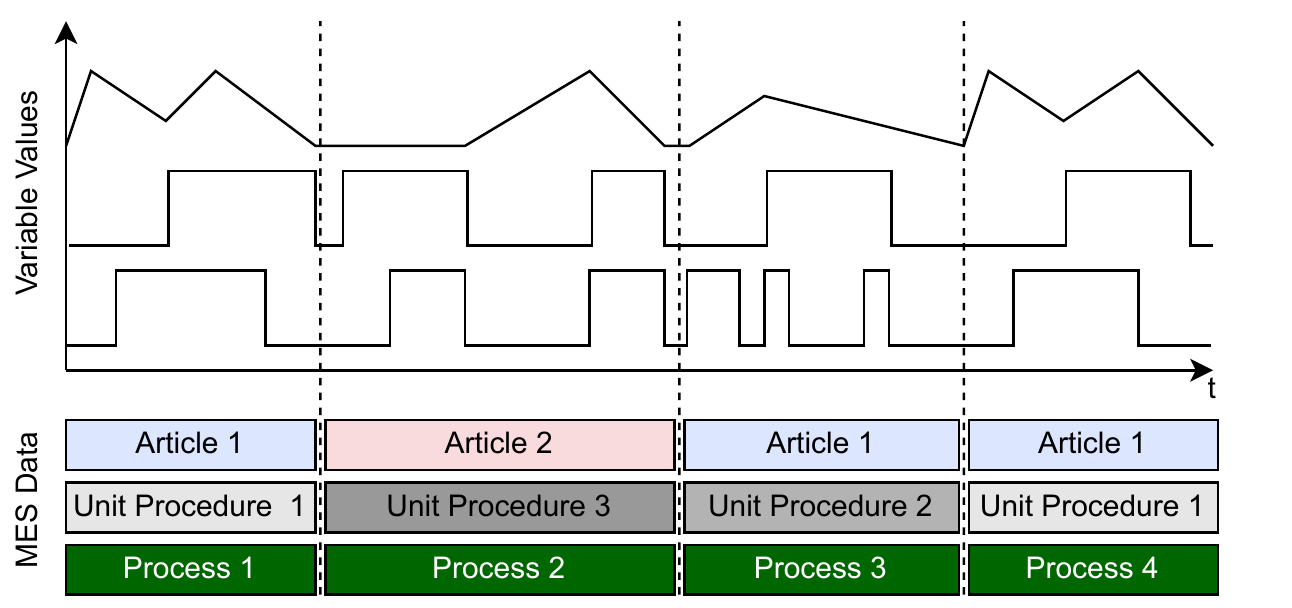}
    \caption{Data access principle for a single machine: Processes can be selected based on various criteria. Using the start and end time of the processes, individual event traces can be extracted.}
    \label{fig:ETFA_Access_Principle_Diagram}
\end{figure}

The knowledge graph was filled with descriptions of processes using an OWL representation of the formalized process description. These descriptions contain information on the technical resource that was assigned to the process, the ISA88-type of process that was executed, as well as its inputs, outputs and start- and end times. This information could usually be found in an MES system. The lower part of Figure \ref{fig:ETFA_Access_Principle_Diagram} shows some of the process information that is available.

\begin{figure}[htp]
    \centering
    \includegraphics[width=0.35\textwidth]{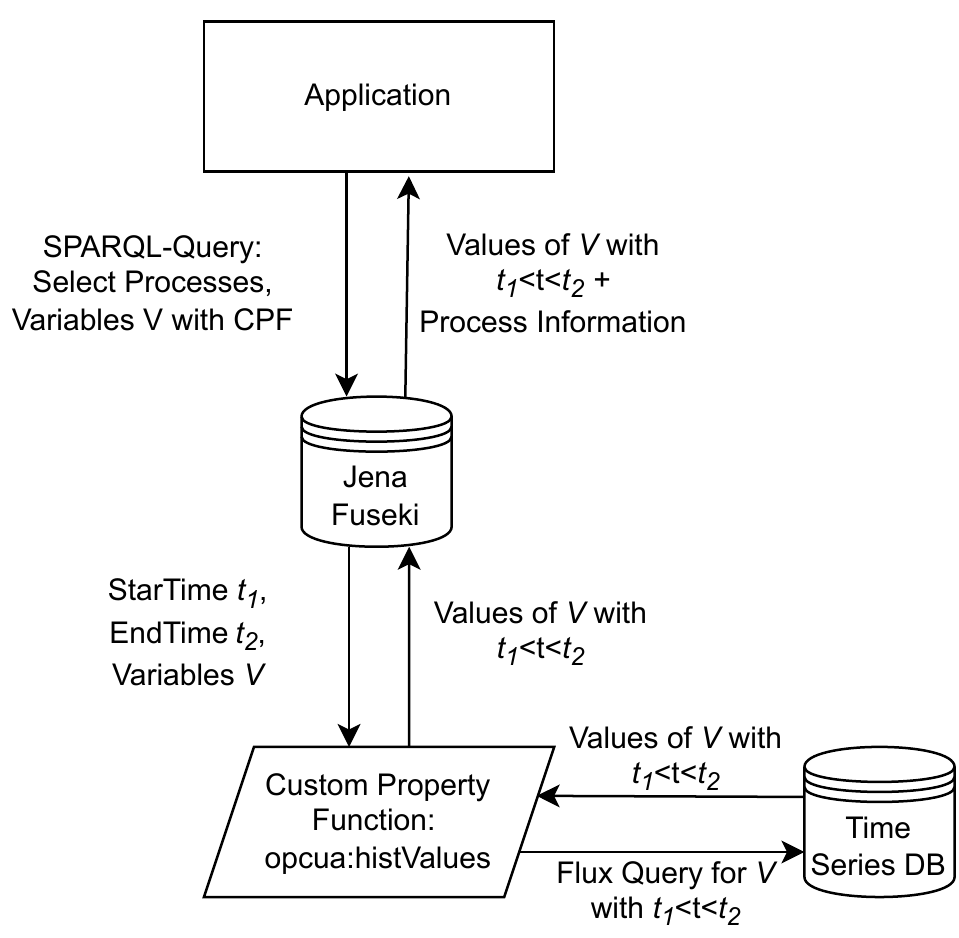}
    \caption{Information flow of a query including the CPF.}
    \label{fig:Query_Workflow}
\end{figure}

Using this information, it is possible to select process instances based on various selection criteria e.g. the article that was produced or the procedure that was realized in this process via SPARQL queries. Since the technical resource of the process is known, the variables of that resource can be extracted from the OPC UA information model. Variables and timestamps are then passed to the Custom Property Function \textit{opcua:histValues()}, which queries the variable values from the time series database. 
Figure \ref{fig:Query_Workflow} describes the information flow for one of these queries.

Technical resource, start time and end time of the process are the only required process description elements to access the recorded OPC UA variable values. 
However, a more granular description of the process (e.g. only processes that were passed a specific parameter as an input) would allow for more precise filtering.

\section{Validation} \label{validation}
In order to assess whether the knowledge graph and the access to the time series database meets the requirements from Section \ref{semanticModel}, SPARQL queries were written to answer the competency questions from Table \ref{tab:RQ1} and Table \ref{tab:RQ2}. 

The first query (s. Listing \ref{list:SPARQLCQ1}) provides a list of all variables, which belong to a certain machine in the ISA88 information model (CQ1.1).

\begin{lstlisting}[label=list:SPARQLCQ1,
    language = SPARQL,
    captionpos=b, frame=single,
    caption={SPARQL query to answer CQ 1.1 and 1.2 for a machine that follows OPC UA for Machine Tools Companion Specifications. Prefix declaration was ommited for brevity.}]
SELECT ?NodeId ?BrowseName ?Time ?Value 
WHERE { 
  # 1. Selecting Machine in OPC UA-Information Model
  ?urnid owl:sameIndividualAs OpcSS:FullMachineTool.
  
  # 2. Selecting Variables from OpcUa-Model
  ?urnid OpcUa:hasComponent* ?node.
  ?node  OpcUa:browseName ?BrowseName.
  ?node  OpcUa:nodeId ?NodeId.
  ?node  OpcUa:typeDefinition ?nodeclass.
  FILTER( ?nodeclass = "BaseDataVariableType" || 
          ?nodeclass = "FiniteStateVariableType" || 
          ?nodeclass = "AnalogUnitRangeType"  ).
          
  # 3. Extracting Variable Values via CPF
  ?node  OpcUa:histValues ( ?Time ?Value 
                            "2022-02-28T09:00:00Z" 
                            "2022-02-28T09:10:00Z").
}
ORDER BY ASC(?time)
\end{lstlisting}

To do this, it navigates the OPC UA information model and extracts all variables whose type description fits the proper categories. If the line \textit{?node OpcUa:histValues (?Time ?Value "starttime" "endtime")} is included, the query will also trigger the custom property function described in Section \ref{useCase}. Once executed, it constructs a query and sends it to InfluxDB to access all variable value changes by any variable that fits the selection criteria defined in the first part of the query. This way, the query also answers CQ1.2.

\begin{table}[h]
\begin{center}
    \begin{tabular}{l c c l} \hline
        \multicolumn{1}{c} {Time} & \multicolumn{1}{c}{Value} & \multicolumn{1}{c}{NodeId} & {BrowseName} \\ \hline
        \rule{0pt}{2ex}2022-02-28T09:00:54Z & true & "ns=7;i=56510" & "IsRotating"  \\ 
        2022-02-28T09:01:26Z & true & "ns=7;i=56519" & "Locked"  \\ 
        2022-02-28T09:01:73Z & false & "ns=7;i=56510" & "IsRotating"  \\ 
        2022-02-28T09:01:70Z & "H²" & "ns=7;i=56600" & "UtilityName"  \\ 
        ... & ... & ... & ... \\ \hline
    \end{tabular}
\end{center}
\caption{Exemplary SPARQL response to CQ 1.2.}
\label{tab:CQ1.2Answer}
\end{table}

It then returns the recorded value changes of all variables along with the timestamps, \textit{NodeIds} and human readable \textit{BrowseName} as shown in Table \ref{tab:CQ1.2Answer}. 
It should be noted that in the example the query accesses events. However, since InfluxDB does not differentiate between time series and events, the approach is equally valid for both types of time based data.

The queries for CQ 2.1 and 2.2 (see Listings \ref{list:SPARQLCQ2} and \ref{list:SPARQLCQ2.2}) build upon the functionality shown in Listing \ref{list:SPARQLCQ1}. 

\begin{lstlisting}[label=list:SPARQLCQ2,
    language = SPARQL,
    captionpos=b, frame=single,
    caption={SPARQL query to answer CQ 2.1 by filtering for Unit Procedure. Prefix declaration was ommited for brevity.}]
SELECT ?Time ?Value ?NodeId ?BrowseName ?Process
WHERE { 
    # 1. selecting ISA88-Procedure and -Unit:
    ?proc    ISA88:isRealizedInProcessStage ?Process.
    ?unit    VDI3682:isAssignedTo ?Process.
    FILTER(  ?proc = OpcSS:UnitProcedure1).
    FILTER(  ?unit = OpcSS:FullMachineTool).
    
    # 2. selecting Timestamps of Processes
    ?Process VDI3682:hasInput ?stimeDE.
    ?stimeDE DINEN61360:hasTypeDescription 
                OpcSS:StartTimeProcess;
            DINEN61360:hasInstanceDescription / 
            DINEN61360:Value ?starttime.
    ?Process VDI3682:hasOutput ?etimeDE.
    ?etimeDE DINEN61360:hasTypeDescription
                OpcSS:EndTimeProcess;
    ?etimeDE DINEN61360:hasInstanceDescription / 
            DINEN61360:Value ?endtime.
    ?Process a VDI3682:Process.
    
    # 3. selecting Variables from OpcUa-Model
    ?urnid   owl:sameIndividualAs ?unit.
    ?urnid   OpcUa:hasComponent* ?node.
    ?node    OpcUa:browseName ?BrowseName;
            OpcUa:typeDefinition ?nodeclass;
            OpcUa:nodeId ?NodeId.
    FILTER(  ?nodeclass = "BaseDataVariableType" || 
            ?nodeclass = "FiniteStateVariableType" ||
            ?nodeclass = "AnalogUnitRangeType").
            
    # 4. Extracting Variable Values via CPF
    ?node    OpcUa:histValues (?Time ?Value 
                                ?starttime ?endtime).
}
ORDER BY ASC(?Time)
\end{lstlisting}

They expanded by a query that extracts start and end timestamps from processes that can then be passed to the \textit{OpcUa:histValues}-function. This way, only variables between the timestamps are returned. 

The query can be adapted to filtering processes by any type of input, output or Technical Resource. This would also allow to filter by different information elements (e.g. process parameters) that function as input or output of the process. 

The ISA88-information model offers another possibility to filter the selected processes by certain conditions with regards to its physical or procedural model. Since the information model contains explicit information about the type of process, it is possible to only select realizations of specific \textit{Unit Procedures} or \textit{Operations}. 

The SPARQL query in Listing \ref{list:SPARQLCQ2} extracts the start and end time from any process that is a realization of a specific \textit{Unit Procedure} and executed by a specific \textit{Unit}. It then retrieves the Units' variables via the OPC UA information model and queries the variable value changes from InfluxDB through the Custom Property Function \textit{OpcUa:histValues()}. 

With minor alterations in the first part, the same query can be used to answer CQ 2.2. As shown in Listing \ref{list:SPARQLCQ2.2}, only three lines in the first part need to be changed , while the remaining part of the query stays the same. 

\begin{lstlisting}[label=list:SPARQLCQ2.2,
    language = SPARQL,
    frame=single, captionpos=b, 
    caption={SPARQL query to answer CQ 2.2 by filtering for Product Type. Query sections 2., 3., and 4. remain the same as in Listing \ref{list:SPARQLCQ2}.}]
SELECT ?Time ?Value ?NodeId ?BrowseName ?Process
WHERE { 
    # 1. selecting Article and -Unit:
    ?proc    ISA88:isRealizedInProcessStage ?Process.
    ?unit    VDI3682:isAssignedTo ?Process.
    ?Process VDI3682:hasOutput ?Product.
    ?Product OpcSS:hasProductType ?article.
    FILTER(  ?article = OpcSS:Article1).
    FILTER(  ?unit = OpcSS:FullMachineTool).
    
    # 2. selecting Timestamps of Processes
    ?Process VDI3682:hasInput ?stimeDE.
    ?stimeDE DINEN61360:hasTypeDescription 
                OpcSS:StartTimeProcess;
            DINEN61360:hasInstanceDescription / 
            DINEN61360:Value ?starttime.
    ?Process VDI3682:hasOutput ?etimeDE.
    ?etimeDE DINEN61360:hasTypeDescription
                OpcSS:EndTimeProcess;
    ?etimeDE DINEN61360:hasInstanceDescription / 
            DINEN61360:Value ?endtime.
    ?Process a VDI3682:Process.
    
    # 3. selecting Variables from OpcUa-Model
    ?urnid   owl:sameIndividualAs ?unit.
    ?urnid   OpcUa:hasComponent* ?node.
    ?node    OpcUa:browseName ?BrowseName;
            OpcUa:typeDefinition ?nodeclass;
            OpcUa:nodeId ?NodeId.
    FILTER(  ?nodeclass = "BaseDataVariableType" || 
            ?nodeclass = "FiniteStateVariableType" ||
            ?nodeclass = "AnalogUnitRangeType").
            
    # 4. Extracting Variable Values via CPF
    ?node    OpcUa:histValues (?Time ?Value 
                                ?starttime ?endtime).
}
ORDER BY ASC(?Time)
\end{lstlisting}

Other filter criteria like selecting all sub processes that belong to a specific recipe are possible as well.  This has the added benefit that filtering can be applied using the ODPs of the Formalized Process Description and ISA88, which are easily understandable for end users.

An example of the SPARQL response to either query can be seen in Table \ref{tab:CQ2Answer}. It contains the value and timestamp of a node along with the process that was executed during the event. 
Grouping the result by process would allow for an easy creation of event traces for a process. 

\addtolength{\tabcolsep}{-3.1pt}
\begin{table}[htbp]
    \begin{center}
        \begin{tabular}{l c c l l} \hline
            \multicolumn{1}{c} {Time} & \multicolumn{1}{c}{Value} & \multicolumn{1}{c}{NodeId} & {BrowseName} & \multicolumn{1}{c}{Process} \\ \hline
            \rule{0pt}{2ex}2022-02-28T09:00:54Z & true & "ns=7;i=56510" & "IsRotating" &  Process 1 \\ 
            2022-02-28T09:01:26Z & true & "ns=7;i=56519" & "Locked" & Process 1 \\ 
            2022-02-28T09:02:73Z & false & "ns=7;i=56510" & "IsRotating" & Process 1 \\ 
            ... & ... & ... & ... & ... \\ 
            2022-02-28T10:03:36Z & true & "ns=7;i=56510" & "IsRotating" & Process 4 \\ 
            2022-02-28T10:04:11Z & true & "ns=7;i=56519" & "Locked" &  Process 4 \\ 
            ... & ... & ... & ... & ... \\ \hline
        \end{tabular}
    \end{center}
    \caption{Exemplary SPARQL response to CQ 2.1 and CQ 2.2.}
    \label{tab:CQ2Answer}
\end{table}

\section{Conclusion}\label{discussion}
In this paper, a semantic model was presented that allows access to all recorded variable value changes from an OPC UA server based on a Formalized Process Description. To achieve this, the variable changes are logged in a time series data base and can then be accessed through a custom property function in a SPARQL query to the graph data base.
Since the knowledge graph contains information about the production assets, the procedures they can perform and the processes that were realized (i.e. from an MES), this allows for the extraction of individual event traces of specific processes. 
The extracted event traces can then be used to analyse the timing information of processes down to the level of individual variable value changes. 
The approach was successfully validated on exemplary competency questions, which were answered by a number of SPARQL queries. 

While the approach can answer the competency questions, some directions for future work remain. 
So far the prior knowledge is modeled based on the ISA 88 industry standard for batch control. The connection between process types, process instances and Technical Resources are described using other terminology in other domains. To broaden the scope of this approach, this information could also be modeled using skill based approaches\cite{Koecher2020} or the information models outlined in the 
VDI 5600 - Manufacturing Execution Systems \cite{VDI5600}.

The knowledge about the processes that were executed are currently materialized in the triple store. At a production facility that has an MES, this kind of data would usually be stored inside a dedicated relational database. To avoid duplication of data, this database could be directly accessed using an OBDA-tool like Ontop.

\section*{Acknowledgment}
This work has been partially supported and funded by the German Federal Ministry of Education and Research (BMBF) for the project ”Time4CPS - A Software Framework for the Analysis of Timing Behaviour of Production and Logistics Processes” under the contract number 01IS20002.
It was partially developed within the Fraunhofer Cluster of Excellence "Cognitive Internet Technologies".

\bibliographystyle{IEEEtran}
\bibliography{references.bib}

\vspace{12pt}

\end{document}